\documentclass[10pt,conference,compsocconf]{IEEEtran}
\usepackage{booktabs}
\usepackage{hyperref}
\usepackage{amssymb}
\usepackage{subcaption}
\usepackage{graphicx}	% For figure environment
\usepackage{url}
\usepackage{amsmath}
\usepackage{tabularx}
\usepackage{pifont}
\usepackage{makecell}

\usepackage{graphicx} % for \resizebox
\usepackage{array} 
\usepackage{caption}
\begin{document}
% Obviamente no es el definitivo
\title{Automatic Segmentation of 3D CT scans with SAM2 using a zero-shot approach}

\author{
  Miquel López Escoriza \& Pau Amargant Alvarez\\
  \textit{Department of Computer Science, EPFL, Switzerland}
}

\maketitle

\begin{abstract}

Foundation models for image segmentation have shown strong generalization in natural images, yet their applicability to 3D medical imaging remains limited. In this work, we study the zero-shot use of Segment Anything Model 2 (SAM2) for automatic segmentation of volumetric CT data, without any fine-tuning or domain-specific training. We analyze how SAM2 should be applied to CT volumes and identify its main limitation: the lack of inherent volumetric awareness. To address this, we propose a set of inference-alone architectural and procedural modifications that adapt SAM2’s video-based memory mechanism to 3D data by treating CT slices as ordered sequences. We conduct a systematic ablation study on a subset of 500 CT scans from the TotalSegmentator dataset to evaluate prompt strategies, memory propagation schemes and multi-pass refinement. Based on these findings, we select the best-performing configuration and report final results on the a bigger sample of the TotalSegmentator dataset comprising 2,500 CT scans. Our results show that, even with frozen weights, SAM2 can produce coherent 3D segmentations when its inference pipeline is carefully structured, demonstrating the feasibility of a fully zero-shot approach for volumetric medical image segmentation.

\end{abstract}

\section{Background}

%\subsection{Supervised Segmentation and Data Constraints}
Medical image segmentation has traditionally relied on fully supervised deep learning. Convolutional architectures like U-Net and V-Net~\cite{ronneberger2015unet, milletari2016vnet} set the standard by effectively capturing local features, while subsequent Vision Transformers (ViTs) (e.g., UNETR) were introduced to model long-range dependencies.
Despite their effectiveness, these models are task-specific and constrained by the "annotation bottleneck": they require large, pixel-perfectly labeled datasets for training and cannot generalize to unseen classes without retraining.

%\subsection{Shift to Interactive Segmentation}
The Segment Anything Model (SAM)~\cite{kirillov2023sam} introduced a shift from task-specific automation to general-purpose interactive segmentation. Trained on broad data, SAM is a foundation model that acts as a class-agnostic tool designed to accelerate annotation via user prompts (points, boxes).
However, SAM is inherently 2D. When applied to 3D medical volumes, it processes slices independently. 

%\subsection{SAM 2 and Volumetric Consistency}
SAM 2~\cite{ravi2024sam2} addresses these temporal limitations by introducing a streaming memory mechanism for video segmentation. By treating a CT or MRI volume as a video, SAM 2 can store features from prompted slices and propagate them to adjacent ones. This theoretically restores 3D spatial consistency while maintaining the flexibility of prompt-based interaction.

%\subsection{Adaptation Strategies}
Most existing adaptations of SAM~2 to medical imaging, such as Medical SAM~2~\cite{zhu2024medicalsam2, ma2024medsam}, address the domain gap through supervised fine-tuning. These approaches typically modify the architecture or retrain model weights on task-specific medical datasets in order to achieve competitive performance.

While effective, fine-tuning introduces several practical limitations. Curated medical datasets with dense, high-quality annotations are often difficult and costly to obtain, and training large foundation models requires substantial computational resources. Moreover, task-specific fine-tuning may compromise the model’s general-purpose capabilities, leading to reduced performance outside the target medical domain.

In contrast, this work focuses on a strictly zero-shot setting. We conduct a systematic ablation study that keeps all SAM~2 weights frozen and evaluates how different inference-time design choices—such as prompting strategies, memory selection, and propagation schemes—affect performance on volumetric medical segmentation. The objective is to assess whether volumetric awareness can emerge from inference configuration alone, without additional training, while preserving the general-purpose nature of the foundation model.

\section{Methods: From Video to Volume}

\begin{figure*}[ht!]
    \centering
    \includegraphics[width=1.0\linewidth]{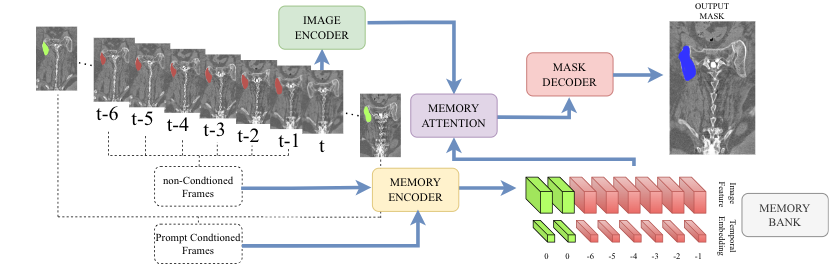}
    \caption{\textbf{Adaptation of SAM2 to 3D CT volumes}. A CT scan is interpreted as a pseudo-video by treating the depth ($z$) axis as the temporal dimension. At each slice $t$, the input image is processed by the image encoder to produce an embedding. Embeddings from prompted slices (first$\&$last frame in the figure) and a fixed number of neighboring slices are stored in the memory bank and augmented with learned temporal embeddings. Prompt-conditioned frames are assigned a zero temporal offset, while non-conditioned frames use learned temporal embeddings. The memory attention module attends to a fixed window of six frames, producing a fused representation that is passed to the mask decoder to generate the segmentation mask for slice $t$.}
    \label{fig:video_to_volume}
\end{figure*}

\subsection{Overview and Problem Formulation}

SAM2 is a promptable segmentation model designed for images and videos. Given one or more frames annotated with prompts, the model predicts segmentation masks and propagates them to subsequent frames by leveraging a memory mechanism. This propagation is achieved through a memory bank that stores intermediate visual embeddings from previously processed frames. As seen in Figure ~\ref{fig:video_to_volume}, when segmenting a new frame, SAM2 attends to this memory bank to maintain consistency of the segmented object over time.

As in prevous approaches \cite{ma2024medsam}, \cite{zhu2024medicalsam2} we adapt SAM2 to volumetric data by reinterpreting a 3D CT volume as a pseudo-video. We map the depth ($z$) axis of the volume to the temporal dimension, treating each axial slice as a video frame. We choose the axial direction because CT scans typically have higher resolution and more consistent spacing along this axis.

Nevertheless, applying SAM2 directly to 3D medical volumes presents a fundamental challenge: maintaining coherence across the depth of the volume. In videos, the relationship between frames is inherently temporal, and the model exploits the assumption that object appearance evolves smoothly over time. In contrast, consecutive slices in a CT volume are related spatially rather than temporally. Anatomical structures do not move or deform over time but are sampled across space, which can result in abrupt appearance changes, partial visibility, or topological variations between neighboring slices.

Figure~\ref{fig:video_to_volume} illustrates the inference pipeline used to adapt SAM2 to volumetric data. Our goal is not to retrain the model, but to evaluate how far its frozen architecture can be pushed toward volumetric awareness through inference-time design alone. We therefore conduct a systematic ablation study that modifies only the inference pipeline, memory usage, and propagation strategy, while keeping all model weights fixed.

\subsection{Dataset and Technical Protocol}

All experiments are conducted on the TotalSegmentator dataset \cite{wasserthal2023totalsegmentator}, using the official high-resolution CT subset with voxel-level annotations for 104 anatomical structures. In this study, we focus exclusively on bone structures, which present a challenging segmentation target due to their complex geometry and frequent similarity to surrounding tissues.

Each sample is treated as a full 3D volume. To separate exploratory analysis from final evaluation, we define two fixed subsets: a subset of 500 CT volumes used for extensive ablation studies, and a larger subset of 2{,}500 CT volumes used to report final performance for the best-performing configurations. All splits are balanced across bone tissue, are immutable and predefined to ensure reproducibility. More information can be found in Appendix \ref{sec:dataset}.

We follow common evaluation protocols in promptable segmentation where user interaction is simulated from ground-truth annotations to ensure reproducibility and reduce prompt variability~\cite{interactive3dmedsam2, espmedsam2024}. 
Specifically, we use \emph{mask prompts} rather than \emph{point and box prompts}.These are derived from the ground truth, which provide a consistent and unambiguous initialization across volumes and methods. Since SAM2 is already reliable for single-slice segmentation when given a strong prompt, this choice is not intended to inflate performance, but to ensure that performance differences primarily reflect the behavior of the propagation and memory mechanisms rather than variability in prompt placement.

As detailed in Appendix \ref{sec:metrics} segmentation performance is evaluated per anatomical structure using the Dice Similarity Coefficient (DSC), Intersection over Union (IoU), and Hausdorff distance.

Inference is memory-bound due to repeated movement of feature maps through the memory bank. While the pipeline can be executed on CPU at reduced speed, GPU acceleration is required for large-scale evaluation. Deployment via ONNX is possible and is a viable path for practical applications. For further information, please refer to Appendix.
\ref{sec:compute}
\subsection{Preprocessing}

CT scans encode tissue density using Hounsfield units spanning a wide intensity range. For bone segmentation, only a restricted interval is informative. We therefore apply intensity windowing tailored to bone tissue, followed by normalization and conversion to three-channel inputs compatible with the SAM2 image encoder.

To further reduce domain mismatch, volumes are spatially cropped to retain only the relevant anatomical region. This assumption is realistic in clinical workflows, where rough localization is readily available.

\subsection{Architectural Tweaks for Volume Awareness}

The core limitation of SAM2 in volumetric settings lies in its memory bank, which treats frames strictly as past temporal events. In 3D volumes, spatial distance rather than temporal recency determines relevance. We introduce several inference-time modifications to make the memory mechanism volume-aware along a primary axis.

\subsubsection{Selection of Conditioned Frames}
By default, all prompted frames are included in the memory bank with equal importance. In CT bone segmentation, this can be counterproductive: visually similar but anatomically distinct structures may dominate attention, and the memory mechanism has no notion of spatial proximity. We therefore explore alternative strategies for selecting which prompted frames are kept in memory.

\subsubsection{Temporal Tracking}
SAM2 attends to a fixed number of recent frames using learned temporal embeddings. We investigate how changing the effective memory size  affects volumetric coherence. We also focus on the learned temporal embeddings and how can one tweak them to make the model volume-aware.

\subsubsection{Multi-Axis Propagation}
\begin{figure}[h!]
    \centering
    \includegraphics[width=1.0\linewidth]{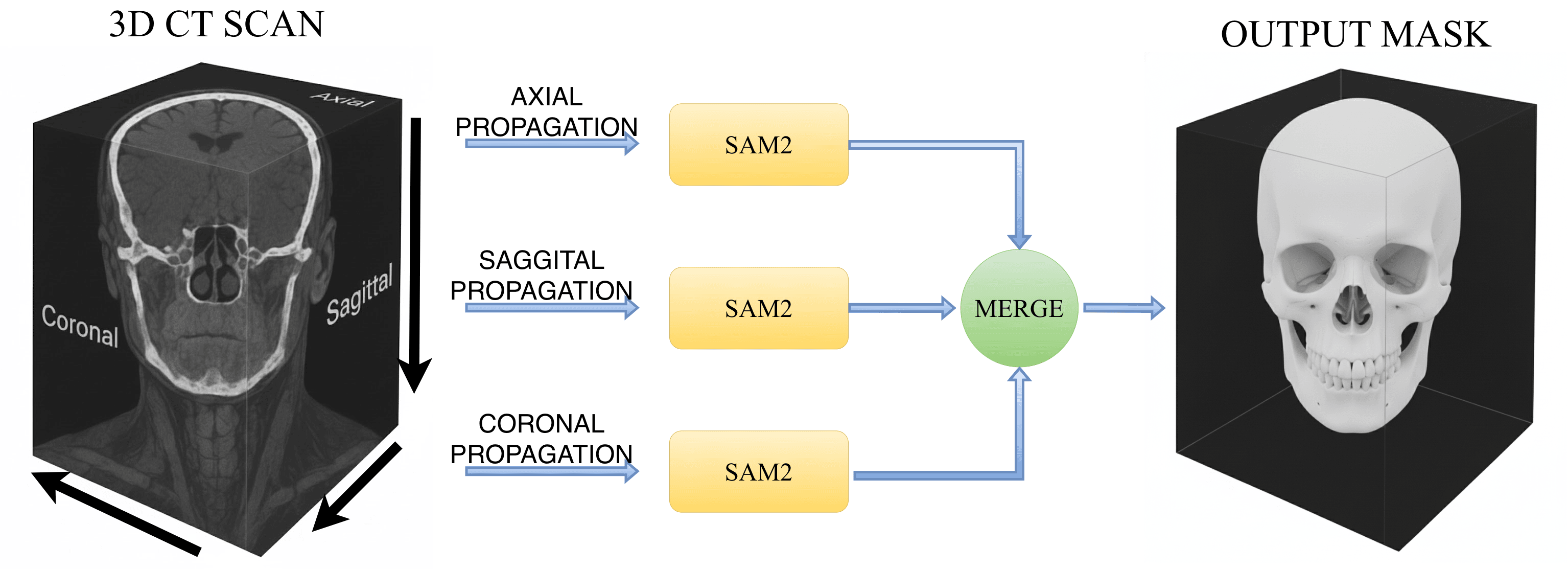}
    \caption{\textbf{Multi-axis propagation and fusion strategy}. A 3D CT volume is segmented independently along the axial, sagittal, and coronal axes using SAM2. Each axis produces a logit volume, which are reoriented to a common reference frame and merged to obtain the final segmentation.}

    \label{fig:THREE_AXIS}
\end{figure}

Volumetric data is inherently isotropic, and restricting inference to a single axis may introduce directional bias. As illustrated in Figure \ref{fig:THREE_AXIS}, we therefore propagate segmentation independently along the axial, coronal, and sagittal axes and fuse the resulting predictions. Let $L_A$, $L_C$, and $L_S$ denote the logit volumes obtained from each axis; the final prediction is computed as
$
P_{\text{fused}} = \sigma\!\left( \frac{L_A + L_C + L_S}{3} \right),
$
where $\sigma$ denotes the sigmoid function. In addition, we evaluate bidirectional (forward and backward) propagation along a single axis to reduce bias induced by a fixed traversal direction.

\section{Results}

In this section, we present the quantitative evaluation of the proposed zero-shot pipeline on the TotalSegmentator dataset. We first analyze the impact of preprocessing, model scale and prompting strategy to establish a stable baseline. Subsequently, we evaluate the contribution of memory and propagation mechanisms.

\begin{figure*}[ht!]
    \centering
    \includegraphics[width=1.0\linewidth]{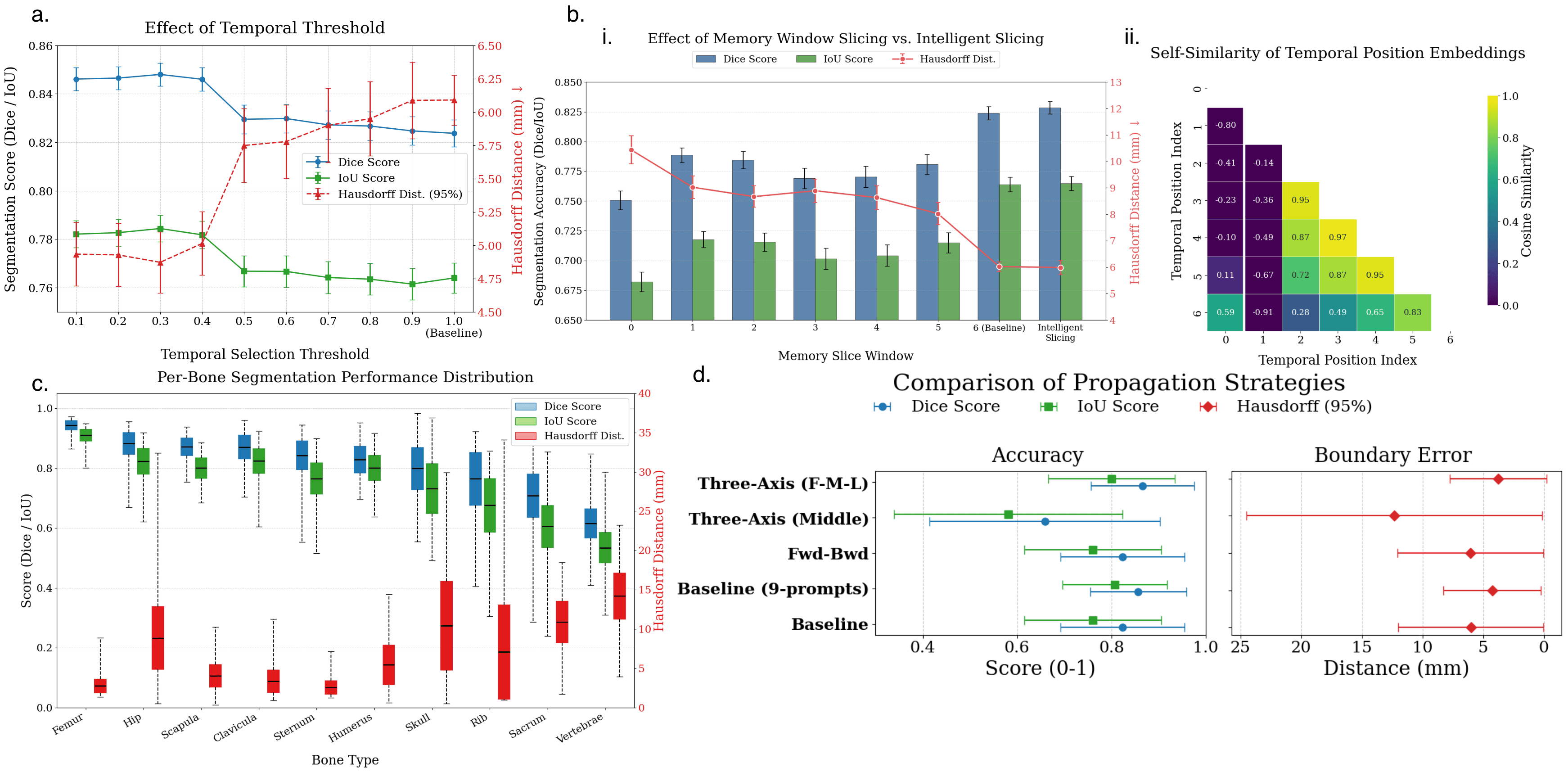}
\caption{\textbf{Ablation study on memory design and propagation strategies.}
(a) Effect of the prompt-conditioned memory threshold, showing improved performance when only nearby prompted slices are retained.
(b) (i) Influence of non-conditioned memory size on performance, and (ii) cosine similarity of temporal embeddings, highlighting redundancy among intermediate past frames. (c) Per-category segmentation performance, illustrating variability across anatomical structures.(d) Performance comparison of different volume propagation strategies under varying prompt budgets, including Forward–Backward and Three-Axis variants.}

    \label{fig:memory_bank}
\end{figure*}

\subsection{Baseline Setting}
We establish our baseline configuration through an ablation study summarizing the impact of model scale, preprocessing, and prompting (Tab.~\ref{tab:baseline_ablation}). We select the \textbf{Small} model variant as it approximates the performance of the Large model with significantly lower computational cost. For preprocessing, \textbf{Windowing Alone} is adopted; it improves segmentation accuracy by \textbf{30\%} compared to raw data, outperforming CLAHE. Finally, the \textbf{First--Middle--Last} strategy is chosen as the prompt baseline, offering the optimal trade-off between segmentation performance and annotation budget. As shown in Fig.~\ref{fig:memory_bank}c, the baseline performance varies across bone tissues, indicating that segmentation difficulty is highly dependent on the specific anatomical structure.

\subsection{Memory Impact}

We first examine the effect of the prompt-conditioned memory bank. This is controlled by a temporal distance threshold that determines whether a prompted slice is retained in memory based on its relative position within the volume. As shown in Fig.~\ref{fig:memory_bank}a, smaller thresholds consistently yield higher performance. Including prompted frames that are far apart is less effective and can introduce noise, since all conditioned frames share the same temporal embedding. An optimal threshold is observed around $0.3$ (which will be later referred to as \emph{Structured Prompt Selection or SPS}), resulting in an improvement of approximately \textbf{2\%} over the baseline.

We next analyze the role of the non-conditioned memory bank, which stores embeddings from recently processed and non-prompted slices. Fig.~\ref{fig:memory_bank}b.i shows that attending to the last six frames—the default SAM2 configuration—achieves the strongest overall performance. The trend is non-linear: performance improves from $0$ to $1$ frame, remains unchanged from $1$ to $5$, and increases again at $6$. The embedding similarity analysis in Fig.~\ref{fig:memory_bank}b.ii provides insight into this behavior, revealing that intermediate temporal embeddings are highly redundant. Consequently, storing only these embeddings contributes little additional information, whereas including both early and late temporal embeddings introduces more distinctive context.

\begin{figure*}
    \centering
    \includegraphics[width=0.9\linewidth]{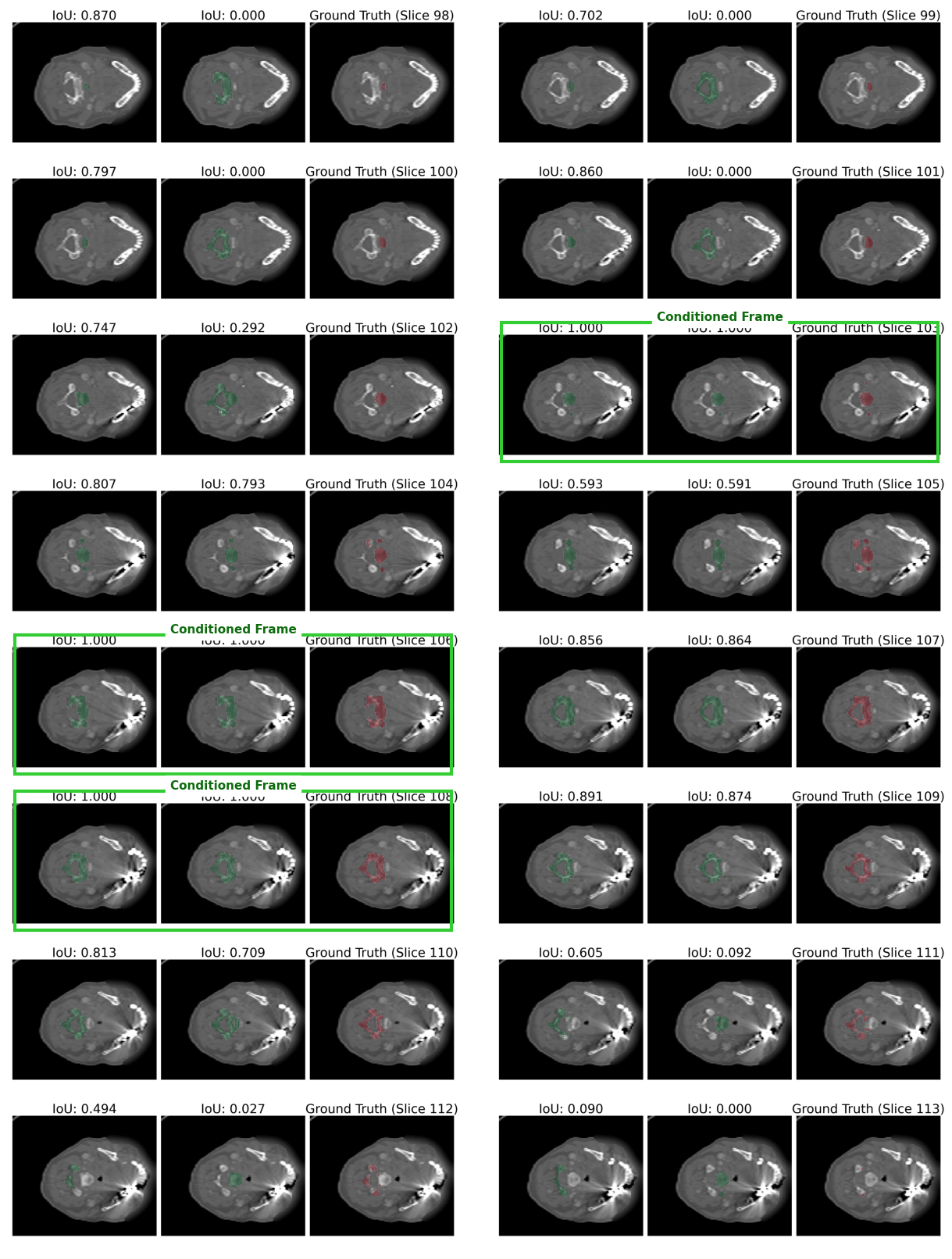}
    \caption{\textbf{Vertebra segmentation produced by the baseline model.} The proposed configuration (IS), and the ground truth. Each triplet corresponds to a slice of the CT volume. Vertebrae exhibit strong slice-to-slice appearance changes, making consistent tracking across the volume challenging. The proposed method produces more consistent segmentations across slices and successfully tracks the correct vertebra throughout the volume. While some predicted masks remain imprecise, the target structure is preserved. In contrast, the baseline model frequently segments incorrect objects (e.g., slices 100 and 111), likely due to confusion with visually similar structures in distant slices that are included in its memory.}

    \label{fig:visualization_example}
\end{figure*}

Motivated by the embedding analysis, we evaluate an \emph{Intelligent Slicing or IS} strategy for the non-conditioned memory bank. We retain only the two most recent frames and assign them the first and last temporal embeddings, which slightly outperforms both the baseline and also the unmodified two-frame variant by approximately \textbf{4\%}. Furthermore, it reduces the tracking runtime of the model as seen in Fig.~\ref{fig:boxplot}. 
This indicates that the opening and closing temporal embeddings are particularly important, and that for 3D CT volumes, attending to fewer, more recent frames is preferable to including distant slices that are likely to differ abruptly as can bee seen in Fig.~\ref{fig:visualization_example}. 
%Figure ~\ref{fig:visualization_example} compares the resulting segmentation using IS and baseline configurations.

We next compare different volume propagation strategies (Fig.~\ref{fig:memory_bank}c). The baseline and Forward--Backward strategies both use three prompts, with Forward--Backward underperforming the baseline. The Three-Axis strategy, using nine prompts (three per axis), achieves the best performance. When constrained to a single prompt per axis (three prompts total), Three-Axis performs significantly worse than the baseline. However, when both methods use nine prompts, Three-Axis again outperforms the baseline, showing that with a larger prompt budget, distributing prompts across axes is more effective than concentrating them along a single direction. 

Tab.~\ref{tab:final_ablation} summarizes the evaluation of the leading strategies on the full 2,500-volume dataset. Combining the best-performing components yields an overall improvement of approximately \textbf{4\%} while also reducing inference runtime. At the anatomical level (see Tab.~\ref{tab:per_bone_results}), vertebrae—one of the most temporally non-smooth structures—benefit from a \textbf{12.5\%} absolute Dice improvement, indicating that the proposed inference design improves the model’s volumetric awareness.

A more detailed ablation is reported in Tab.~\ref{tab:advanced_ablation}, where we further study memory strides and confidence-based gating. Increasing the stride modestly improves performance by expanding the spatial coverage of the memory bank. In contrast, gating strategies that retain only the top-$k$ most confident frames degrade performance. This is likely because high-confidence frames cluster near prompted slices, which are already present in memory, resulting in redundancy rather than additional context.

%Report that for vertebrae is 10 percent increase on performance. 
  
\begin{table}[t]
\centering
\footnotesize
\setlength{\tabcolsep}{4pt}
\begin{tabular}{lccc}
\toprule
\textbf{Method} & \textbf{Dice $\uparrow$} & \textbf{IoU $\uparrow$} & \textbf{HD $\downarrow$} \\
\midrule
NP & 0.594 $\pm$ 0.242 & 0.509 $\pm$ 0.227 & 11.177 $\pm$ 9.366 \\
Baseline & 0.804 $\pm$ 0.134 & 0.734 $\pm$ 0.148 & 6.240 $\pm$ 6.563 \\
SPS & 0.823 $\pm$ 0.117 & 0.751 $\pm$ 0.134 & 5.236 $\pm$ 5.740 \\
IS & 0.822 $\pm$ 0.125 & 0.759 $\pm$ 0.138 & 5.683 $\pm$ 5.178 \\
\textbf{IS + SPS} & \textbf{0.841 $\pm$ 0.107} & \textbf{0.778 $\pm$ 0.122} & \textbf{4.788 $\pm$ 4.930} \\
\bottomrule
\end{tabular}
\caption{\textbf{Evaluation on the full dataset (2,500 volumes).} \\
NP refers to \emph{No Preprocessing}, SPS to \emph{Structured Prompt Selection} (temporal threshold of 0.3), and IS denotes \emph{Intelligent Slicing}.}
\vspace{-20pt}
\label{tab:final_ablation}
\end{table}

\section{Conclusion}

In this work, we investigated how far SAM2 can be applied to 3D CT segmentation in a strictly zero-shot setting. By treating CT volumes as ordered slice sequences, we analyzed the mismatch between SAM2’s video-based assumptions and the spatial characteristics of volumetric medical data. Through a systematic ablation study, we showed that inference-time design choices—particularly preprocessing, prompt placement, memory selection, and propagation strategy—have a strong impact on performance, even when model weights are frozen.

Our results highlight several practical insights. Segmentation difficulty varies across anatomical structures, especially for tissues that exhibit abrupt spatial changes and violate assumptions of smooth temporal evolution. Prompt-conditioned memory entries that are spatially distant are often detrimental. Temporal embeddings play a central role, with the inclusion of both starting and ending temporal embeddings being particularly important. In addition, attending to fewer, more recent frames—rather than maintaining a large memory bank—often yields better results for CT volumes. When a larger prompt budget is available, distributing prompts across multiple axes improves segmentation quality, albeit at increased computational cost. However, not all video-inspired strategies transfer well: confidence-based gating, for example, tends to introduce redundancy rather than meaningful additional context.

Overall, this study clarifies what can and cannot be achieved with SAM2 through inference-time modifications alone. While the proposed strategies enable coherent 3D segmentations without training, performance remains constrained by the model’s original design. Further improvements will likely require architectural changes or medical-domain adaptation.

\newpage

\section*{Ethical Risks}

This project studies the use of a foundation model (SAM2) for zero-shot segmentation of 3D medical images. The main ethical risk relates to the potential misuse of the produced segmentations in clinical contexts. The primary stakeholders affected would be patients and clinicians. If segmentation outputs were used directly for diagnosis or treatment planning without proper validation, errors could lead to incorrect medical decisions. While such consequences could be severe, the likelihood is low within the scope of this work, as the system is intended purely for research and is not deployed in any clinical setting.

We evaluated this risk by analyzing segmentation performance across multiple anatomical structures using quantitative metrics (Dice, IoU, and Hausdorff distance). The observed variability in performance highlights that the method is not uniformly reliable and reinforces that it should not be considered clinically ready. These results informed how we present the work and its limitations.

To mitigate this risk, we clearly frame the project as an academic study of inference-time behavior, not as a medical tool. All experiments are conducted on a publicly available and fully anonymized dataset, ensuring that no personal or sensitive patient data is exposed. We also avoid any claims about diagnostic accuracy or clinical applicability and restrict conclusions to methodological insights.

A secondary ethical consideration is computational and environmental cost. By keeping all model weights frozen and avoiding additional training, the project limits energy consumption compared to fine-tuning large models. Overall, the identified ethical risks are acknowledged and addressed through careful framing, dataset choice, and conservative interpretation of results.

\newpage
\appendices

\section{ML4Science}\label{sec:project_background}
This Project arises from a collaboration with Jonathan Dong and Pablo Garcia-Amorena from the Biomedical Imaging Laboratory at EPFL through EPFL's CS433 ML4Science projects. Their goal was to evalaute whether generalists video segmentation models could be effectively used for CT segmentation, with a particular interest in how SAM 2 would fit within an interactive annotation pipeline where a medical professional manually segments a few prompts and the model extends it to the whole volume. 

In addition, there was interest in evaluating how SAM 2's memory functions and the impact that it has on the model's computational requriements and run-time.

\section{Protocol}\label{sec:protocol}

\subsection{Dataset Construction} \label{sec:dataset}

We use the the Total Segmentator dataset \cite{wasserthal2023totalsegmentator}. The dataset is composed of 1228 CT scans, each with up to 80 anatomical structures annotate. We select only those annotation that correspond to bones. From this we create two evaluation datasets from which we report the ablation study. The first dataset is composed of 500 samples and is balanced to have the same number of each bone type. A second dataset of 2,500 samples is created to report the final results. The compositions of the whole dataset and the two subsets are reported in tables \ref{tab:raw_counts}, \ref{tab:final_dataset} and \ref{tab:aggregation}
.The dataset is generated in the following way:
\begin{enumerate}
    \item We select the annotated structures that correspond to bones and create a count of each bone type.  Similar structures such as different vertebrae, ribs or right/left pair of bones are treated as the same bone.
    \item For  each bone type, segmentations are sampled until the target number is desired. All bone types occur in the same frequency in the final dataset.  Only segmentations that contian more than six axial slices are used.
\end{enumerate}

\subsection{Metrics} \label{sec:metrics}

Segmentation performance is evaluated per anatomical structure using the Dice Similarity Coefficient (DSC), Intersection over Union (IoU), and Hausdorff distance. Dice and IoU measure volumetric overlap and are defined as
\[
\mathrm{Dice} = \frac{2 |P \cap G|}{|P| + |G|}, \qquad
\mathrm{IoU} = \frac{|P \cap G|}{|P \cup G|}.
\]
Boundary accuracy is assessed using the Hausdorff distance,
\[
\mathrm{HD}(P,G) = \max\!\left(
\sup_{p \in P} \inf_{g \in G} \|p-g\|,
\sup_{g \in G} \inf_{p \in P} \|g-p\|
\right).
\]
Where $P$ denotes the predicted segmentation and $G$ the corresponding ground-truth mask.

\subsection{Implementation Details and Computation  Resources} \label{sec:compute}

All experiments were conducted using two compute environments to accommodate the significant graphical memory requirements of the SAM 2 video pipeline. The ablation studies were initially performed utilizing a single NVIDIA RTX 3090 (24 GB VRAM). During some parts of the study an additional NVIDIA A100 GPU was utilized to accelerate the throughput.

The computational resources were a constraint during the execution of the baseline study and limited the number of samples that could be included in the evaluation datasets. Each evaluation of the 500 sample dataset required approximately 50 minutes of run-time while the larger dataset required around 10 hours of run-time.

\subsection{Configuration impact on inference speed} \label{sec:runtime}
In addition to considering the segmentation metrics, we performed an analysis to determine how the time that SAM 2 spends to produce a segmentation is spent and how different configurations impact it. 

\begin{figure*}[ht!]
    \centering
    % First Image
    \begin{subfigure}[b]{0.48\linewidth}
        \centering
        \includegraphics[width=\linewidth]{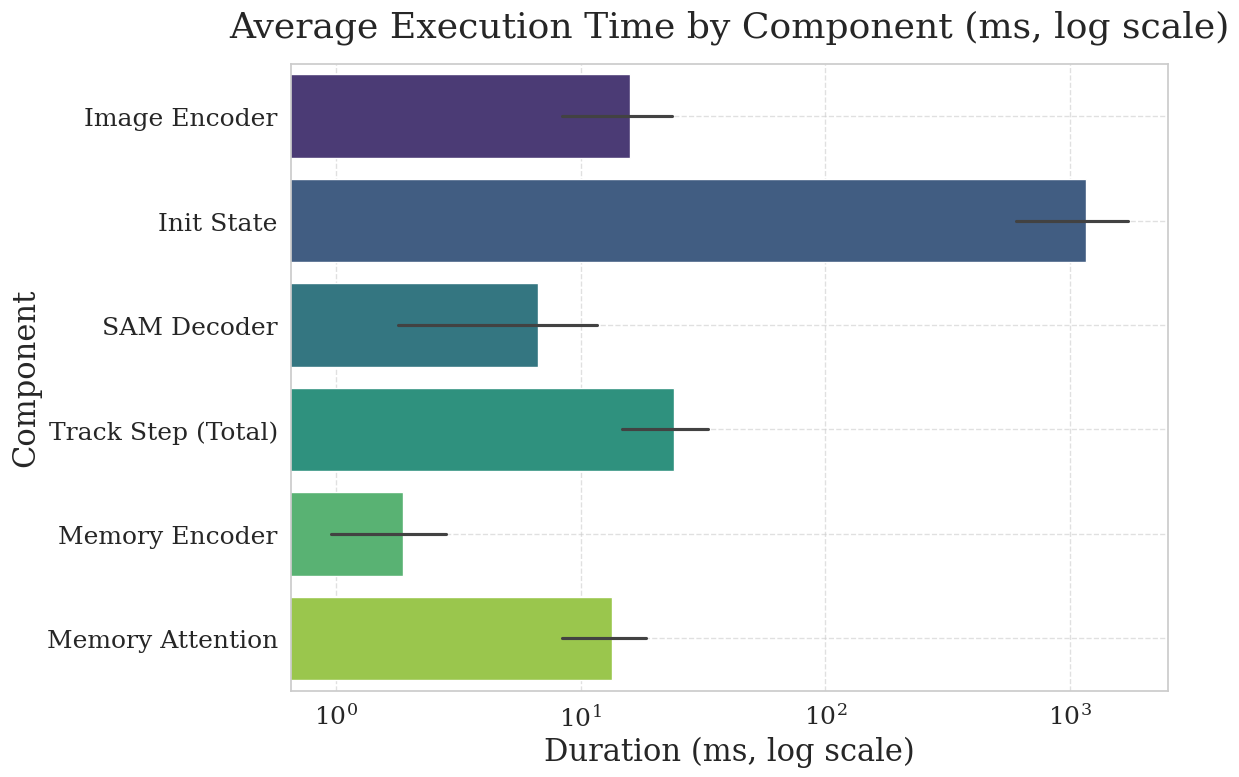}
        \caption{Bar plot showing the average execution time per frame of the main processes that SAM2 performs to predict a mask. }
        \label{fig:barplot}
    \end{subfigure}
    \hfill % Adds spacing between the two images
    % Second Image
    \begin{subfigure}[b]{0.48\linewidth}
        \centering
        \includegraphics[width=\linewidth]{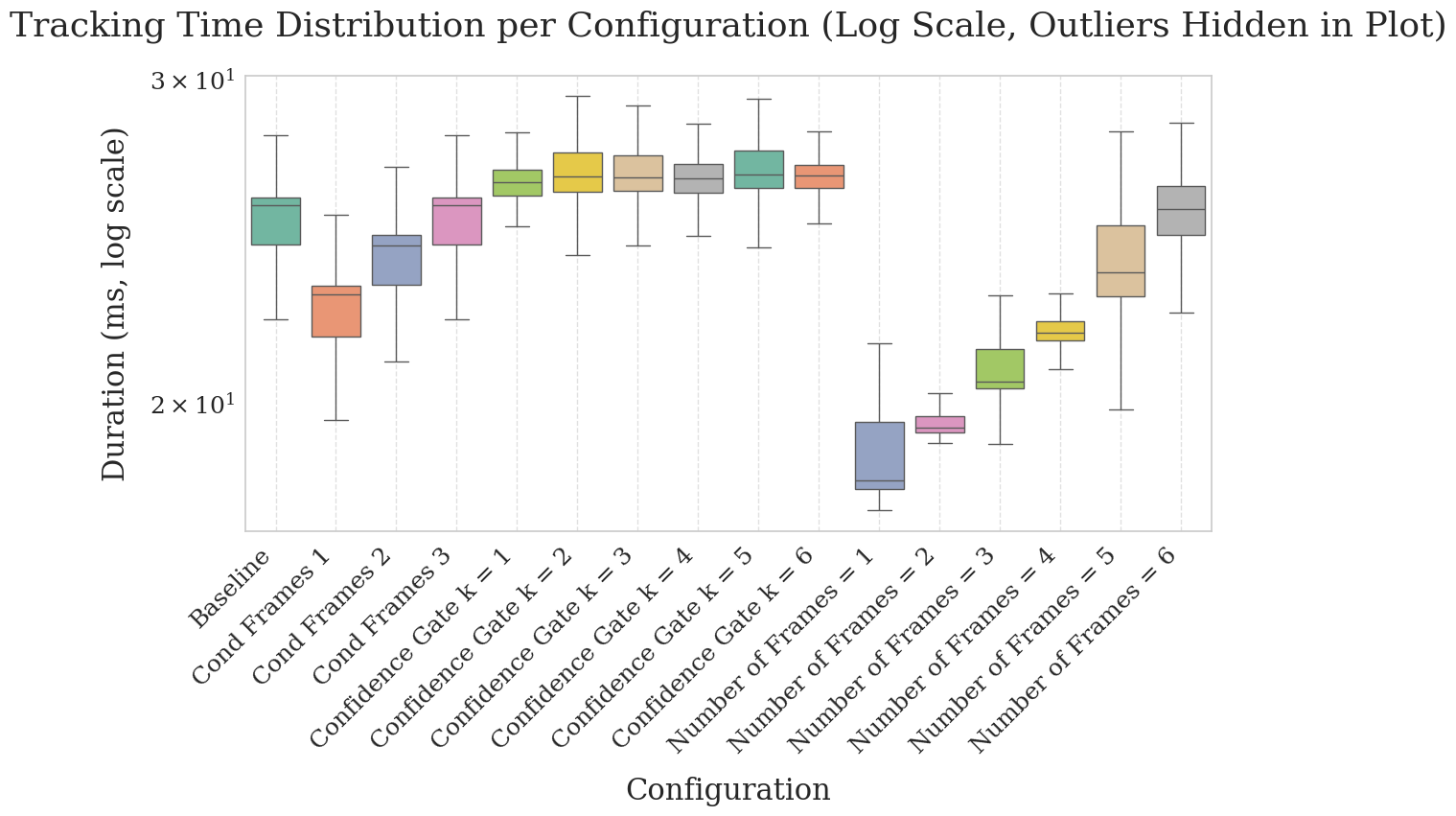}
        \caption{Boxplot showing the tracking time distribution across differnet configurations (log-scale).}
        \label{fig:boxplot}
    \end{subfigure}
    
    \caption{Comparative Runtime analysis across different configurations. Evaluation was performed on a set of 50 bones.}
    \label{fig:runtime_analysis}
\end{figure*}

Figure \ref{fig:barplot} shows the average execution time of the different components that are executed as part of the SAM2 inference pipeline. The state initialization process, which calculates the image embeddings across all frames dominates the run-time while the memory encoder and memory attention processes have run-times limited impact on the execution time.

Similarly, an analysis was performed to determine whether the modifications in the model's memory had an impact on the execution time required to predict the masks for a frame. Figure \ref{fig:boxplot} shows that including more frames in the memory has a clear impact in the tracking execution time.

% ==========================================
% TABLE 1: Raw Data Distribution
% ==========================================
\begin{table*}[t]
\centering
\small
\renewcommand{\arraystretch}{1.2}
\setlength{\tabcolsep}{0pt}
\captionsetup{font=small, labelfont=normalfont, textfont=normalfont}

\begin{tabular*}{\textwidth}{@{\extracolsep{\fill}} l l c c}
\hline
\textbf{Anatomical Structure} & \textbf{Specific Labels} & \textbf{Count (Each)} & \textbf{Total Raw Masks} \\
\hline
Ribs       & \texttt{rib\_left\_1..12}, \texttt{rib\_right\_1..12} & 1,228 & 29,472 \\
Vertebrae  & \texttt{C1..7}, \texttt{T1..12}, \texttt{L1..5}, \texttt{S1} & 1,228 & 30,700 \\
Femur     & \texttt{femur\_left}, \texttt{femur\_right} & 1,228 & 2,456 \\
Humerus   & \texttt{humerus\_left}, \texttt{humerus\_right} & 1,228 & 2,456 \\
Clavicle  & \texttt{clavicula\_left}, \texttt{clavicula\_right} & 1,228 & 2,456 \\
Scapula   & \texttt{scapula\_left}, \texttt{scapula\_right} & 1,228 & 2,456 \\
Hip       & \texttt{hip\_left}, \texttt{hip\_right} & 1,228 & 2,456 \\
Sacrum    & \texttt{sacrum} & 1,228 & 1,228 \\
Sternum   & \texttt{sternum} & 1,228 & 1,228 \\
Skull     & \texttt{skull} & 1,228 & 1,228 \\
\hline
\end{tabular*}

\vspace{6pt}
\caption{\textbf{Raw Data Distribution (Pre-Consolidation).} Availability of anatomical structures in the TotalSegmentator dataset. Note that all selected structures possess complete annotation coverage across the 1,228 scans.}
\label{tab:raw_counts}
\end{table*}

% ==========================================
% TABLE 2: Aggregation Logic
% Spans full width (table*) to handle long math expressions
% ==========================================
\begin{table*}[t]
\centering
\small
\renewcommand{\arraystretch}{1.2}
\setlength{\tabcolsep}{0pt}
\captionsetup{font=small, labelfont=normalfont, textfont=normalfont}

\begin{tabular*}{\textwidth}{@{\extracolsep{\fill}} l l l}
\hline
\textbf{Target Class} & \textbf{Aggregation Type} & \textbf{Constituent Raw Labels} \\
\hline
Vertebrae & Serial Union & $\bigcup (\texttt{C1..7}, \texttt{T1..12}, \texttt{L1..5}, \texttt{S1})$ \\
Rib       & Serial \& Bilateral & $\bigcup_{i=1}^{12} (\texttt{rib\_left\_i} \cup \texttt{rib\_right\_i})$ \\
Hip       & Bilateral Union & $\texttt{hip\_left} \cup \texttt{hip\_right}$ \\
Humerus  & Bilateral Union & $\texttt{humerus\_left} \cup \texttt{humerus\_right}$ \\
Femur    & Bilateral Union & $\texttt{femur\_left} \cup \texttt{femur\_right}$ \\
Clavicle & Bilateral Union & $\texttt{clavicula\_left} \cup \texttt{clavicula\_right}$ \\
Scapula  & Bilateral Union & $\texttt{scapula\_left} \cup \texttt{scapula\_right}$ \\
Sacrum   & Identity & $\texttt{sacrum}$ \\
Sternum  & Identity & $\texttt{sternum}$ \\
Skull    & Identity & $\texttt{skull}$ \\
\hline
\end{tabular*}

\vspace{6pt}
\caption{\textbf{Class Aggregation Mapping.} The mapping function $f$ used to consolidate raw labels into the final ten bone classes. The aggregation logic accounts for lateral symmetry (left/right) and serial structures.}
\label{tab:aggregation}
\end{table*}

% ==========================================
% TABLE 3: Final Ablation Dataset
% Standard table, centered.
% ==========================================
% ==========================================
% TABLE 3: Final Ablation Dataset
% ==========================================
\begin{table*}[t]
\centering
\small
\renewcommand{\arraystretch}{1.2}
\setlength{\tabcolsep}{0pt}
\captionsetup{font=small, labelfont=normalfont, textfont=normalfont}

\begin{tabular*}{\textwidth}{@{\extracolsep{\fill}} l c c c}
\hline
\textbf{Bone Class} & \textbf{Consolidated Total Available} & \textbf{Ablation Subset ($N$)} & \textbf{Final Evaluation Subset ($N$)} \\
\hline
Vertebrae & 30,700 & 50 & 250 \\
Rib       & 29,472 & 50 & 250 \\
Humerus  & 2,456  & 50 & 250 \\
Clavicle & 2,456  & 50 & 250 \\
Hip      & 2,456  & 50 & 250 \\
Scapula  & 2,456  & 50 & 250 \\
Femur    & 2,456  & 50 & 250 \\
Sacrum   & 1,228  & 50 & 250 \\
Sternum  & 1,228  & 50 & 250 \\
Skull    & 1,228  & 50 & 250 \\
\hline
\textbf{Total} & \textbf{76,136} & \textbf{500} & \textbf{2,500} \\
\hline
\end{tabular*}

\vspace{6pt}
\caption{\textbf{Final Balanced Dataset for Ablation.} The subset of 500 volumes curated for the ablation study, ensuring equal representation ($N=50$) across the ten aggregated classes.}
\label{tab:final_dataset}
\end{table*}

\begin{table*}[t]
\centering
\small
\renewcommand{\arraystretch}{1.2}
\setlength{\tabcolsep}{0pt}

% This command removes bolding from "Table 1" and the text
% Requires \usepackage{caption} in your preamble
\captionsetup{font=small, labelfont=normalfont, textfont=normalfont}

\begin{tabular*}{\textwidth}{@{\extracolsep{\fill}} l l l c c c}
\hline
\textbf{Model Scale} & \textbf{Preprocessing} & \textbf{Prompt Strategy} & \textbf{IoU} & \textbf{Dice} & \textbf{Hausdorff} \\
\hline
\multicolumn{6}{l}{\textit{Experiment 1: Varying Model Scale}} \\
Tiny            & Windowing Alone & F--M--L & $0.7402 \pm 0.1632$ & $0.8040 \pm 0.1528$ & $6.78 \pm 7.09$ \\
Base            & Windowing Alone & F--M--L & $0.7478 \pm 0.1357$ & $0.8133 \pm 0.1200$ & $6.92 \pm 6.17$ \\
\textbf{Small}  & \textbf{Windowing Alone} & \textbf{F--M--L} & $\mathbf{0.7604 \pm 0.1448}$ & $\mathbf{0.8236 \pm 0.1314}$ & $\mathbf{6.02 \pm 5.97}$ \\
Large           & Windowing Alone & F--M--L & $0.7638 \pm 0.1332$ & $0.8291 \pm 0.1169$ & $5.80 \pm 5.61$ \\
\hline
\multicolumn{6}{l}{\textit{Experiment 2: Varying Preprocessing}} \\
Small & No preprocessing       & F--M--L & $0.4672 \pm 0.2625$ & $0.5411 \pm 0.2852$ & $11.5853 \pm 9.8032$ \\
Small & CLAHE + Windowing & F--M--L & $0.7535 \pm 0.1460$ & $0.8193 \pm 0.1311$ & $6.08 \pm 6.37$ \\
Small & CLAHE Alone        & F--M--L & $0.7414 \pm 0.1457$ & $0.8113 \pm 0.1313$ & $6.54 \pm 6.54$ \\
Small & CLAHE + Windowing & F--M--L & $0.7535 \pm 0.1460$ & $0.8193 \pm 0.1311$ & $6.08 \pm 6.37$ \\

\textbf{Small} & \textbf{Windowing Alone} & \textbf{F--M--L} & $\mathbf{0.7604 \pm 0.1448}$ & $\mathbf{0.8236 \pm 0.1314}$ & $\mathbf{6.02 \pm 5.97}$ \\
\hline
\multicolumn{6}{l}{\textit{Experiment 3: Varying Prompt Strategy}} \\
Small & Windowing Alone & Middle (1)      & $0.5536 \pm 0.2257$ & $0.6086 \pm 0.2343$ & $15.98 \pm 14.94$ \\
Small & Windowing Alone & First--Last (2) & $0.7214 \pm 0.1912$ & $0.7889 \pm 0.1766$ & $7.69 \pm 9.37$ \\
\textbf{Small} & \textbf{Windowing Alone} & \textbf{F--M--L (3)}     & $\mathbf{0.7604 \pm 0.1448}$ & $\mathbf{0.8236 \pm 0.1314}$ & $\mathbf{6.02 \pm 5.97}$ \\
Small & Windowing Alone & Uniform (5)     & $0.7586 \pm 0.1484$ & $0.8217 \pm 0.1357$ & $6.01 \pm 5.97$ \\
Small & Windowing Alone & Uniform (7)     & $0.7831 \pm 0.1287$ & $0.8395 \pm 0.1173$ & $4.98 \pm 4.69$ \\
\hline
\end{tabular*}
\vspace{6pt}
\captionsetup{name=Tab., labelsep=period, font=small, labelfont=normalfont, textfont=normalfont}

\caption{\textbf{Ablation study establishing the baseline configuration}. We isolate the impact of each component by varying one parameter at a time while fixing the others to their default settings. The \textbf{Small} model is selected as the baseline because it approximates the performance of the Large variant with significantly lower computational cost. For preprocessing, we compare standard intensity windowing against Contrast-Limited Adaptive Histogram Equalization (CLAHE); results show that \textbf{Windowing Alone} yields superior segmentation, improving performance by \textbf{30\%} compared to no preprocessing. Finally, the \textbf{First--Middle--Last} strategy is adopted as the prompt baseline, as it provides the most effective trade-off between performance and annotation budget compared to denser prompting strategies.}
\label{tab:baseline_ablation}
\end{table*}

\begin{table*}[t] 
\centering
\begin{tabularx}{\textwidth}{@{\extracolsep{\fill}} l l c c}
\hline
\textbf{Category} & \textbf{Variant} & \textbf{IoU} & \textbf{Dice} \\
\hline
\multicolumn{4}{l}{\textit{Experiment 1: Memory Bank Temporal Stride}} \\
Memory Bank & Stride 1 (Baseline) & $0.7604 \pm 0.1450$ & $0.8236 \pm 0.1315$
\\
Memory Bank & Stride 2 & $0.7556 \pm 0.1412$ & $0.8189 \pm 0.1267$ \\
\textbf{Memory Bank} & \textbf{Stride 4} & $\mathbf{0.7635 \pm 0.1352}$ & $\mathbf{0.8271 \pm 0.1190}$ \\
\hline
\multicolumn{4}{l}{\textit{Experiment 2: Confidence Gating Strategies}} \\
Gating & Gate 6 & $0.7443 \pm 0.1441$ & $0.8084 \pm 0.1302$ \\
Gating & Gate 5 & $0.7509 \pm 0.1563$ & $0.8135 \pm 0.1456$ \\
Gating & Gate 4 & $0.7540 \pm 0.1528$ & $0.8171 \pm 0.1411$ \\
Gating & Gate 3 & $0.7540 \pm 0.1528$ & $0.8171 \pm 0.1411$ \\
Gating & Gate 2 & $0.7593 \pm 0.1484$ & $0.8227 \pm 0.1354$ \\
\textbf{Gating} & \textbf{Gate 1} & $\mathbf{0.7611 \pm 0.1487}$ & $\mathbf{0.8244 \pm 0.1355}$ \\
Gating & Gate 0 (Baseline) & $0.7604 \pm 0.1450$ & $0.8236 \pm 0.1315$ \\
\hline
\multicolumn{4}{l}{\textit{Experiment 3: Memory Bank Capacity (Slicing)}} \\
Memory Bank & Sliced 0 & $0.6820 \pm 0.1836$ & $0.7507 \pm 0.1743$ \\
Memory Bank & Sliced 1 & $0.7176 \pm 0.1511$ & $0.7887 \pm 0.1361$ \\
Memory Bank & Sliced 2 & $0.7155 \pm 0.1726$ & $0.7845 \pm 0.1638$ \\
Memory Bank & Sliced 3 & $0.7015 \pm 0.1960$ & $0.7692 \pm 0.1923$ \\
Memory Bank & Sliced 4 & $0.7042 \pm 0.2005$ & $0.7705 \pm 0.1971$ \\
Memory Bank & Sliced 5 & $0.7604 \pm 0.1450$ & $0.8236 \pm 0.1315$ \\
Memory Bank & Sliced 6 (Baseline) & $0.7604 \pm 0.1450$ & $0.8236 \pm 0.1315$ \\
\textbf{Memory Bank} & \textbf{Inteligent Slicing } & $\mathbf{0.7780 \pm 0.063}$ & $\mathbf{0.8546 \pm 0.0506}$ \\
\hline
\multicolumn{4}{l}{\textit{Experiment 4: Propagation Strategy}} \\
Propagation, 3 prompts & Three Axis & $0.5816 \pm 0.2424$ & $0.6592 \pm 0.2446$ \\
Propagation, 3 prompts & Forward-Backward & $0.7601 \pm 0.1452$ & $0.8233 \pm 0.1317$ \\
\textbf{Propagation, 3 prompts} & \textbf{Single Axis (Baseline)} & $\mathbf{0.7604 \pm 0.1450}$ & $\mathbf{0.8236 \pm 0.1315}$ \\
\textbf{Propagation, 9 prompts} & \textbf{Three-axis} &$\mathbf{0.8000 \pm 0.1350}$ & $\mathbf{0.8658 \pm 0.1100}$  \\
Propagation, 9 prompts & Single Axis (Baseline) & $0.8071 \pm 0.1115$ & $ 0.8569 \pm 0.1018$ \\

\hline
\multicolumn{4}{l}{\textit{Experiment 5: Temporal Threshold (Temperature)}} \\
Temporal Threshold & Temp 0.1 & $0.7821 \pm 0.1232$ & $0.8461 \pm 0.1057$ \\
Temporal Threshold & Temp 0.2 & $0.7827 \pm 0.1229$ & $0.8465 \pm 0.1054$ \\
\textbf{Temporal Threshold} & \textbf{Temp 0.3} & $\mathbf{0.7844 \pm 0.1228}$ & $\mathbf{0.8480 \pm 0.1052}$ \\
Temporal Threshold & Temp 0.4 & $0.7818 \pm 0.1249$ & $0.8460 \pm 0.1069$ \\
Temporal Threshold & Temp 0.5 & $0.7668 \pm 0.1431$ & $0.8295 \pm 0.1303$ \\
Temporal Threshold & Temp 0.6 & $0.7667 \pm 0.1428$ & $0.8298 \pm 0.1291$ \\
Temporal Threshold & Temp 0.7 & $0.7642 \pm 0.1446$ & $0.8272 \pm 0.1311$ \\
Temporal Threshold & Temp 0.8 & $0.7635 \pm 0.1435$ & $0.8267 \pm 0.1301$ \\
Temporal Threshold & Temp 0.9 & $0.7615 \pm 0.1441$ & $0.8247 \pm 0.1305$ \\
Temporal Threshold & Temp 1.0 (Baseline) & $0.7604 \pm 0.1450$ & $0.8236 \pm 0.1315$  \\
\hline
\hline
\end{tabularx}

\vspace{6pt}
\captionsetup{name=Tab., labelsep=period, font=small, labelfont=normalfont, textfont=normalfont}
\caption{\textbf{Comprehensive ablation study on inference design choices.}
Stride-based memory sampling shows a small benefit at moderate strides (Stride~4). 
Confidence gating does not improve performance and generally degrades results compared to the baseline. 
Reducing memory capacity(without changing temporal embeddings) hurts performance, while intelligent slicing provides the best trade-off.
Multi-axis propagation is only beneficial when using a larger prompt budget, and a moderate temporal threshold ($\text{Temp}=0.3$) yields the best performance.
}
\label{tab:advanced_ablation}
\end{table*}

\begin{table*}[t]
\centering
\small
\renewcommand{\arraystretch}{1.2}
\setlength{\tabcolsep}{0pt}
\captionsetup{font=small, labelfont=normalfont, textfont=normalfont}

\begin{tabular*}{\textwidth}{@{\extracolsep{\fill}} l c c c c c c}
\hline
\textbf{Bone Class} 
& \textbf{NP} 
& \textbf{Base} 
& \textbf{SPS} 
& \textbf{IS}
& \textbf{IS + SPS} 
& \textbf{$\Delta$ Dice} \\
\hline
Vertebrae 
& $0.469 \,/\, 0.391$
& $0.609 \,/\, 0.534$
& $0.699 \,/\, 0.614$
& $0.635 \,/\, 0.563$
& $\mathbf{0.734 \,/\, 0.656}$
& $+12.5\%$ \\

Rib   
& $0.432 \,/\, 0.346$
& $0.751 \,/\, 0.648$
& $0.773 \,/\, 0.670$
& $0.783 \,/\, 0.694$ % Rib not provided in raw list
& $\mathbf{0.794 \,/\, 0.708}$
& $+4.3\%$ \\

Humerus  
& $0.775 \,/\, 0.713$
& $0.865 \,/\, 0.831$
& $0.875 \,/\, 0.841$
& $0.858 \,/\, 0.826$
& $\mathbf{0.865 \,/\, 0.834}$
& $+0.0\%$ \\

Clavicle 
& $0.620 \,/\, 0.535$
& $0.847 \,/\, 0.790$
& $0.866 \,/\, 0.808$
& $0.870 \,/\, 0.826$
& $\mathbf{0.882 \,/\, 0.837}$
& $+3.5\%$ \\

Hip      
& $0.633 \,/\, 0.529$
& $0.862 \,/\, 0.796$
& $0.871 \,/\, 0.807$
& $0.909 \,/\, 0.858$
& $\mathbf{0.917 \,/\, 0.863}$
& $+5.5\%$ \\

Scapula  
& $0.553 \,/\, 0.463$
& $0.848 \,/\, 0.769$
& $0.856 \,/\, 0.778$
& $0.866 \,/\, 0.796$
& $\mathbf{0.870 \,/\, 0.803}$
& $+2.2\%$ \\

Femur    
& $0.847 \,/\, 0.774$
& $0.949 \,/\, 0.917$
& $0.950 \,/\, 0.919$
& $0.945 \,/\, 0.915$
& $\mathbf{0.947 \,/\, 0.917}$
& $-0.2\%$ \\

Sacrum   
& $0.422 \,/\, 0.330$
& $0.694 \,/\, 0.585$
& $0.715 \,/\, 0.607$
& $0.724 \,/\, 0.624$
& $\mathbf{0.760 \,/\, 0.658}$
& $+6.6\%$ \\

Sternum  
& $0.600 \,/\, 0.502$
& $0.825 \,/\, 0.743$
& $0.818 \,/\, 0.733$
& $0.824 \,/\, 0.745$
& $\mathbf{0.818 \,/\, 0.737}$
& $-0.7\%$ \\

Skull    
& $0.593 \,/\, 0.515$
& $0.797 \,/\, 0.730$
& $0.802 \,/\, 0.735$
& $0.811 \,/\, 0.747$
& $\mathbf{0.820 \,/\, 0.752}$
& $+2.3\%$ \\
\hline
\end{tabular*}

\vspace{6pt}
\caption{\textbf{Per-bone segmentation performance on the final evaluation set (2,500 volumes).}
Results are reported as Dice / IoU. NP: No Preprocessing; IS: Inteligent Slicing (proposed); SPS: Structured Prompt Selection. 
$\Delta$ Dice indicates absolute change of IS + SPS relative to Base.}
\label{tab:per_bone_results}
\end{table*}

\newpage
\begin{thebibliography}{9}

% --- Traditional 3D Segmentation ---

\bibitem{ronneberger2015unet}
\label{unet}
O. Ronneberger, P. Fischer, and T. Brox, ``U-Net: Convolutional Networks for Biomedical Image Segmentation,'' in \textit{Medical Image Computing and Computer-Assisted Intervention (MICCAI)}, 2015, pp. 234--241.

\bibitem{milletari2016vnet}
\label{vnet}
F. Milletari, N. Navab, and S. A. Ahmadi, ``V-Net: Fully Convolutional Neural Networks for Volumetric Medical Image Segmentation,'' in \textit{Fourth International Conference on 3D Vision (3DV)}, 2016, pp. 565--571.

\bibitem{isensee2021nnunet}
\label{nnunet}
F. Isensee, P. F. Jaeger, S. A. A. Kohl, \textit{et al.}, ``nnU-Net: a self-configuring method for deep learning-based biomedical image segmentation,'' \textit{Nature Methods}, vol. 18, pp. 203--211, 2021.

% --- Foundation Models (SAM & SAM 2) ---

\bibitem{kirillov2023sam}
\label{sam}
A. Kirillov, E. Mintun, N. Ravi, \textit{et al.}, ``Segment Anything,'' in \textit{Proceedings of the IEEE/CVF International Conference on Computer Vision (ICCV)}, 2023, pp. 4015--4026.

\bibitem{ravi2024sam2}
\label{sam2}
N. Ravi, V. Gabeur, Y. T. Hu, \textit{et al.}, ``SAM 2: Segment Anything in Images and Videos,'' \textit{arXiv preprint arXiv:2408.00714}, 2024.

% --- SAM in Medical Imaging ---

\bibitem{ma2024medsam}
\label{samMed}
J. Ma, Y. He, F. Li, \textit{et al.}, ``Segment Anything in Medical Images,'' \textit{Nature Communications}, vol. 15, no. 654, 2024.

\bibitem{zhu2024medicalsam2}
\label{MedSam2}
J. Zhu, A. Hamdi, Y. Qi, \textit{et al.}, ``Medical SAM 2: Segment Medical Images as Video via Segment Anything Model 2,'' \textit{arXiv preprint arXiv:2408.00874}, 2024.

\bibitem{yang2024sam23dmed}
\label{samto3dmed}
Y. Yang, L. Xu, and L. Tian, ``SAM2-3dMed: Empowering SAM2 for 3D Medical Image Segmentation,'' \textit{arXiv preprint arXiv:2412.00042}, 2024.


\bibitem{wasserthal2023totalsegmentator}
\label{totalsegmentator}
J.~Wasserthal \emph{et al.},
``TotalSegmentator: Robust Segmentation of 104 Anatomic Structures in CT Images,''
\textit{Radiology: Artificial Intelligence}, vol.~5, no.~5, 2023.

\bibitem{espmedsam2024}
Y.~Zhang, H.~Chen, and D.~Shen,
``ESP-MedSAM: Efficient and Stable Prompting for Medical Image Segmentation,''
\textit{arXiv preprint arXiv:2406.XXXXX}, 2024.

\bibitem{interactive3dmedsam2}
J.~Zhu, A.~Hamdi, Y.~Qi, and M.~R.~Sabuncu,
``Medical SAM 2: Segment Medical Images as Video via Segment Anything Model 2,''
\textit{arXiv preprint arXiv:2408.00874}, 2024.

\end{thebibliography}
\end{document}